# An Adaptive Genetic Algorithm for Solving N-Queens Problem

**Uddalok Sarkar[1], *Sayan Nag[1]**

[1]Department of Electrical Engineering
Jadavpur University
Kolkata, India
uddaloksarkar@gmail.com, *nagsayan112358@gmail.com

***Abstract*--In this paper a Meta-heuristic approach for solving the *N*-Queens Problem is introduced to find the best possible solution in a reasonable amount of time. Genetic Algorithm is used with a novel fitness function as the Meta-heuristic. The aim of *N*-Queens Problem is to place *N* queens on an *N* x *N* chessboard, in a way so that no queen is in conflict with the others. Chromosome representation and genetic operations like Mutation and Crossover are described in detail. Results show that this approach yields promising and satisfactory results in less time compared to that obtained from the previous approaches for several large values of *N*.**

## 1. Introduction:

Optimization problems in diverse domains including Engineering Design, Structural Optimization, Economics and Scheduling Assignments need mathematical models and Objective Function. These problems can be of two types: Unconstrained (without constraints) and Constrained (with Constraints). They involve both continuous as well as discrete variables. The usual non-linear nature of these problems in association with a handful of restraints being active at the global optima make the task of finding the optimal solutions all the more troublesome. Optimization techniques and approaches have been successfully applied to various systems, namely, Stochastic Systems, Trajectory Systems, and Deterministic Systems. Stochastic Systems comprises of the optimal control problem in the presence of uncertainty. Trajectory Systems involves aerial paths but is mostly restricted to the computation of optimum trajectories of the rockets and space flight trajectories. Deterministic Systems includes the computations of the performance of several mathematical models of diverse engineering design problems for various physical systems modelling with germane objective functions. Traditional methods which are used previously to solve these Multi-objective Optimization problems include Gradient Descent, Dynamic Programming and Newton Methods. But they are not devoid of shortcomings- they are computationally less efficient. The meta-heuristic algorithms came into existence. These meta-heuristic approximate algorithms tend to provide better solutions in a reasonable amount of time. There exists a number of meta-heuristics like Genetic Algorithm (GA) [1], Particle Swarm Optimization (PSO) [2], Gravitational Search Algorithm (GSA) [3], Ant Colony Optimization (ACO) [4,5], Stimulated Annealing (SA) [6,7], Plant Propagation Algorithm (PPA) [8,9] and so on [10,11].

NP (nondeterministic polynomial time) class of problems are those each of which is solvable in polynomial time by a non-deterministic Turing machine. In other words, problems having no deterministic solutions which run in polynomial time are called NP-class problems. They cannot be solved in real time using deterministic approaches because of their high complexity of the order of $2^n$ or ($n$!). Heuristic methods are used to solve these kinds of problems in real time.

*Corresponding Author, Email: nagsayan112358@gmail.com

*N* Queens Problem is such an optimization problem attributed to the class of NP-Complete Problems. The goal of *N* Queens Problem is to suitably place *N* number of Queens on an *N* x *N* chessboard in a way that there is no conflict between them due to the arrangement, that is, there is no intersection between them vertically or horizontally or diagonally. Eight Queens Problem is a just a special case of *N* Queens Problem, where *N* is equal to 8. Jordan B. and Brett S. [12] considered extensions of the *N* Queens Problem including various board topologies and dimensions at the same time surveying already known solutions for the *N* Queens Problem of placing *N* Queens on an *N* x *N* chessboard. They simultaneously provided a simple solution for finding the intersections of diagonals. They explored diverse areas of the problem, asserting various existing and novel hypotheses. Salabat K. et. al. [13] applied Ant Colony Optimization for solving the 8 Queen Problem thereby sowing that ACO can yield better solution in decent amount of time. Farhad S. et. al. [14] devised a novel technique for solving the *N* Queen Problem by using graph theoretical approach- a hybrid of Depth First Traversal and Breadth First Traversal techniques.

Heuristic approaches are required to solve *N* Queens Problem in real time with optimal solutions. Genetic algorithms (GA) is one such powerful heuristic method which is capable of efficiently solve the problem in real time by virtue of its extensively developed exploration and exploitation properties. In this paper we propose a modified and state-of-the-art Genetic Algorithm with a novel, efficient and robust Fitness Function for generating the best solution for *N* Queens Problem with different random initial candidate solutions and calculating and comparing the fitness values for each of the trial solutions. The paper is organized as follows: Section 2 contains a brief description of the *N*-Queens Problem, Section 3 contains a detailed description of the Genetic Algorithm used along with the fitness function, crossover and mutation operators used. Section 4 contains the Results and Table showing the optimal solutions. Conclusion is provided in Section 5.

## 2. N-Queens Problem:

In 1848, A German Chess player Max Bezzel composed the 8-Queens Problem which aims to place 8 Queens in the chess board in such a way that no two Queens can attack each other. In 1850 Franz Nauck gave the 1$^{st}$ solution to this problem and generalized the problem to *N*-Queen problem for *N* non- attacking Queens on an *N* x *N* Chessboard. Time complexity of an *N*-Queen problem is $O(n!)$. Here, we are proposing a heuristic approach to obtain the best solutions for this problem. We are depicting all the arrangements of an *N* x *N* board as an *N*-tuple (*c*1, *c*2, *c*3… *cN*), where ci represents the position of the queen to be in $i^{th}$ column and $c^{th}$ row. Fig.1 shows an arrangement of 8 x 8 chessboard and its 8-tuple representation.

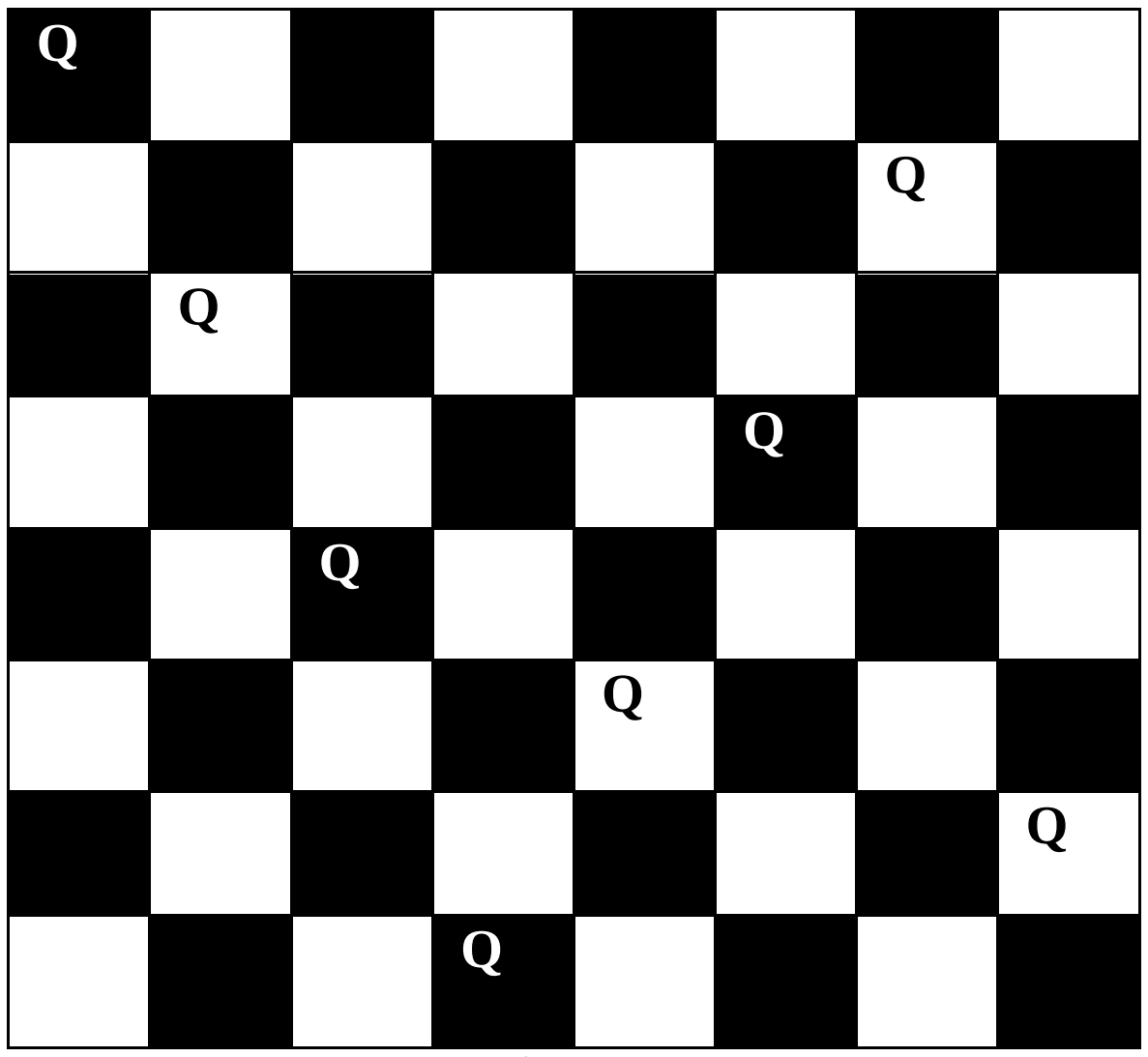

**Fig.1:** Arrangement signifying tuple: (8, 6, 4, 1, 3, 5, 7, 2)

## 3. Genetic Algorithm:

Genetic Algorithm is a widely accepted popular Evolutionary Optimization technique that imitates the process of natural selection. Firstly, in Genetic Algorithm a population of individual chromosomes is selected randomly. These chromosomes are better known as phenotypes in terms of genetics each of which has its individual set of characteristics or traits widely known as genotypes. These genotypes can be altered by adopting various strategies including mutation and crossover. These chromosomes are evolved towards better (eventually optimal) solutions for a combinatorial optimization problem. This optimal solution is dependent on the termination criterion. Genetic Algorithm gives the optimal solution depending on the nature of the fitness function and also on the structure of the algorithm used. Here, Genetic Algorithm is used to solve the $N$ Queens Problem. A general outline how Genetic Algorithm (GA) works is given below:

1. A random population of candidate solutions is created and the fitness scores of the individuals are calculated and the chromosomes are sorted in the population and ranked according to the fitness values.

2. Certain number of chromosomes will pass onto the next generation depending on a selection operator, favoring the better individuals based on their ranking in the population.

3. Selected chromosomes acting as parents will take part in crossover operation to create children whose fitness values are to be calculated simultaneously. Crossover probability is generally kept high, because it is seen to give better children in terms of fitness values.

4. Then based on a mutation probability, a mutation operator is applied on new individuals which randomly changes few chromosomes. Mutation probability is generally kept low.

5. Evaluated off-springs, together with their parents form the population for the next generation.

Steps 2-5 are repeated until a given number of iterations, that is, generations have been evaluated or solution improvement rate falls below some threshold, that is, the difference between two best solutions from consecutive generations is below a certain tolerance limit. Below mentioned are some detailed explanation of the population initialization, the fitness function and the crossover and mutation operators used.

- **Population Initialization:** Firstly we initialize a random population of chromosome of length 1000. Every chromosome here is actually a vector of length $N$, which actually is a random permutation of (1, 2, 3… $N$).

- **Fitness Function:** We design the Fitness of each chromosome in such a way that k number of queens on the same diagonal situation will accumulate $k$-1 points to the fitness value. All these points are summed which obtains the fitness value.

  Let's take the $N$ x $N$ chess board arrangement ($c1$, $c2$, $c3$… $cN$). We'll first check whether more than one Queen is on the same (↗) directed diagonals. This happens if,

  $$\text{For } i \in \{1,2 \ldots N\} \text{ and } j \in \{1,2 \ldots N\} \text{ and } i \neq j;$$

  $$(\mathrm{c}i - i) == (\mathrm{c}j - j)$$

  Similarly, for our next checking whether more than one Queen is on the same (↖) directed diagonals, this happens if,

  $$\text{For } i \in \{1,2 \ldots N\} \text{ and } j \in \{1,2 \ldots N\} \text{ and } i \neq j;$$

$$(ci + i) == (cj + j)$$

Thus we amass the fitness value if any of the above situation occurs. Which clearly shows that the fittest chromosome is signified as 0 fitness value.

The fig.2 and fig.3 shows an ordinary and an optimal arrangement respectively with corresponding fitness values.

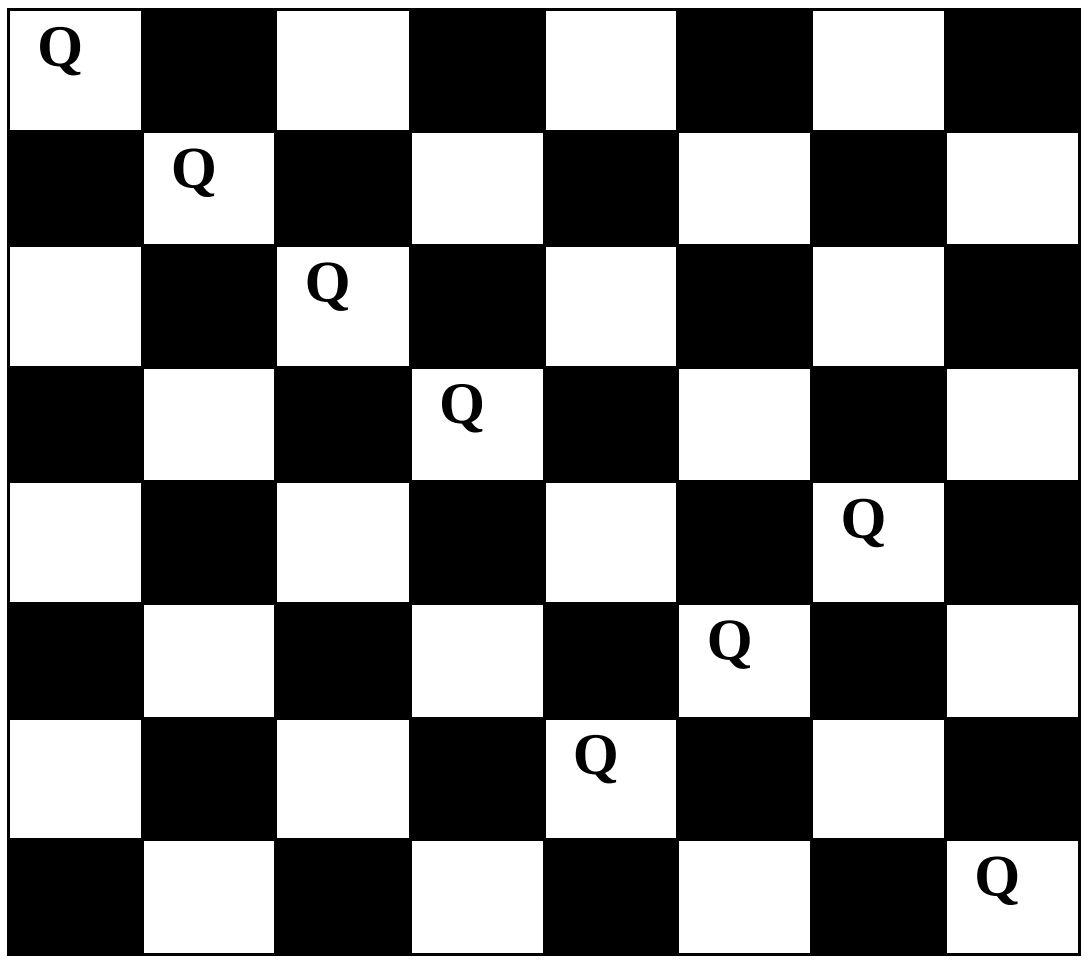

**Fig.2:** Fitness Value: 7

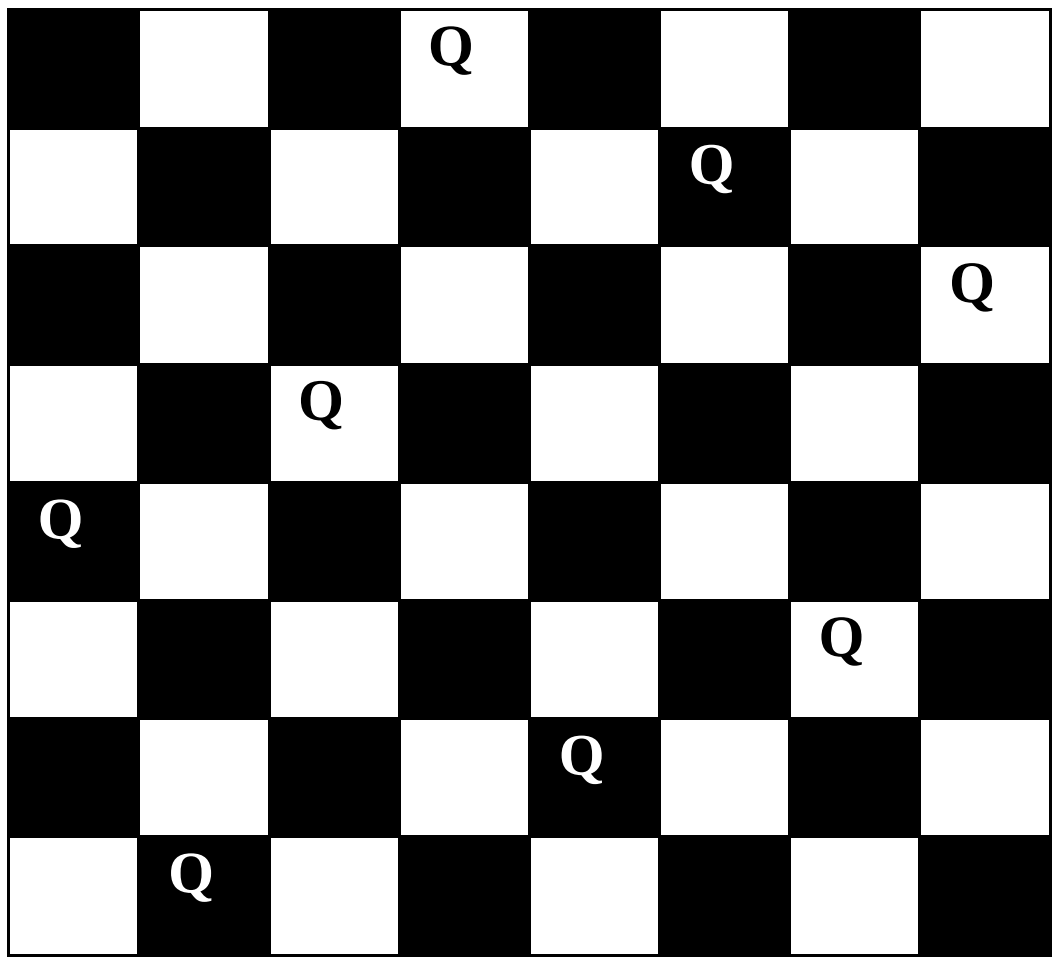

**Fig.3:** Fitness Value: 0

***<u>Pseudo code:</u>***

***Function fitness (*****chromosome*****) {***

```
t1 = 0;   //number of repetitive queens in one diagonal while seen from left corner
t2 = 0;   //number of repetitive queens in one diagonal while seen from right corner
size = length (chromosome);
for i= 1 to size:
   f1 (i) = (chromosome (i)-i);
   f2 (i) = ((1+size)-chromosome (i)-i);
end
f1=sort (f1);
f2=sort (f2);
for i=2 to size:
   if(f1 (i) == f1 (i-1)) //checks whether two Queens are in same diagonals seeing from left corner or not
      t1 = t1+1;
   end
   if(f2 (i) == f2(i-1)) //checks whether two Queens are in same diagonals seeing from right corner or not
      t2 = t2+1;
   end
end
fitness_value = t1 + t2;
return  fitness_value;
}
```

- **Crossover Function:** For obtaining global optimum for a non-convex function the cross over function performs a very important role for obtaining better offspring. As we are considering each of the chromosomes actually a random permutation of (1, 2, 3… *N*), so it was needed to design a permutation crossover. Here we are using the order 1 crossover.

Order 1 Crossover is one sort of very simple permutation crossover in which 2 random points are selected from parent-1 and the alleles between these points are carried over to the child and the other alleles from parent-2 which are absent in child are carried and placed in the child in the order which they appear in parent-2.

Parent1: 5 2 3 1 6 4 8 7
Parent2: ~~1~~ 8 ~~6~~ ~~4~~ 7 5 ~~3~~ 2

Child: 8 7 3 1 6 4 5 2

Step1: In the above scenario, the allele points 3 and 4 are selected randomly from the parent-1 and the alleles between these two points are carried over to the Child chromosome.

Step2: The alleles equivalent to (or equal to) the alleles between the crossover points of parent-1 are eliminated from parent-2.

Step3: The rest of the alleles of parent-2 are placed in the child chromosome according to their order.

Actually we so designed our Crossover Function that it generates two child from the parent chromosome at the same time. Let the parent chromosomes are *A* and *B*. Then the Child-1 is obtained by having *A* as Parent-1 and *B* as Parent-2 and Child-2 is obtained by having *B* as Parent-1 and *A* as Parent-2.

- **Bad Population repository:** We Have made a repository of weak chromosomes. The repository is a size of $\lfloor\sqrt{N}\rfloor$, where $\lfloor x \rfloor$ defines maximum integer less or equal to $x$.

1. In every iteration any 2 chromosomes are selected randomly and gone through crossover and mutation and the offspring competes with the chromosomes of main population. If these offspring are fitter than the worst chromosomes of the main population then the offspring substitutes the worst chromosomes of main population.

2. In every iteration one chromosome is selected randomly and crossed over with any randomly selected chromosome from main population. This offspring also competes with the chromosomes of main population for its existence.

This idea helps us to improve the avoidance of local optimum. In every run we are keeping this repository constant. No updates of chromosome is occurring here.

- **Mutation:** Mutation is very important in genetic algorithm for not to stuck the process in local optimum. Here we have taken mutation probability 0.8. We have also applied double mutation with a probability of 0.4.

Before mutation: 2 6 8 3 4 1 7 5
After mutation: 2 6 7 3 4 1 8 5

In a single mutation function we are randomly selecting 2 alleles and swap them. By running this single mutation two times in a succession we achieve the double mutation function.

## 4. Experimental Results:

Table 1 shows the obtained optimal solutions to *N*-Queens problem for *N* = 10, 11… 25. We have obtained 3 optimal solutions for each *N* using our metaheuristic algorithm. For each run the number of iterations taken by our algorithm to reach the global optimum is also been shown accordingly. The obtained solutions i.e. the elements of *N*-tuples are shown as table format under the Optimal Solution column. Results clearly shows that the optimal results are obtained with much lesser value of iterations.

**Table 1: Optimal Solutions and iterations taken**

| *N* | Iterations | Optimal Solution | | | | | | | | | | | | | | | | | | | | | | | | |
|---|---|---|---|---|---|---|---|---|---|---|---|---|---|---|---|---|---|---|---|---|---|---|---|---|---|---|
| 25 | 1431 | 22 | 6 | 11 | 14 | 24 | 15 | 19 | 8 | 4 | 17 | 1 | 3 | 12 | 16 | 7 | 21 | 18 | 2 | 25 | 23 | 10 | 20 | 5 | 9 | 13 |
| | 382 | 12 | 4 | 16 | 23 | 11 | 24 | 22 | 20 | 2 | 5 | 15 | 10 | 7 | 19 | 3 | 8 | 17 | 25 | 6 | 1 | 18 | 13 | 9 | 14 | 21 |
| | 281 | 17 | 5 | 22 | 8 | 12 | 21 | 1 | 13 | 20 | 16 | 19 | 24 | 11 | 23 | 4 | 7 | 3 | 15 | 9 | 2 | 14 | 10 | 18 | 25 | 6 |
| 24 | 4043 | 6 | 8 | 15 | 17 | 14 | 10 | 23 | 18 | 24 | 2 | 4 | 11 | 7 | 3 | 16 | 13 | 19 | 9 | 22 | 5 | 21 | 12 | 1 | 20 | |
| | 392 | 13 | 3 | 22 | 6 | 18 | 2 | 23 | 11 | 4 | 14 | 17 | 19 | 24 | 8 | 20 | 5 | 1 | 15 | 10 | 7 | 21 | 12 | 9 | 16 | |
| | 420 | 6 | 14 | 21 | 17 | 24 | 16 | 1 | 7 | 4 | 10 | 3 | 18 | 15 | 23 | 11 | 19 | 2 | 5 | 8 | 13 | 22 | 20 | 9 | 12 | |
| 23 | 1363 | 5 | 14 | 22 | 3 | 8 | 13 | 21 | 4 | 6 | 23 | 19 | 2 | 18 | 7 | 17 | 11 | 9 | 16 | 10 | 20 | 15 | 1 | 12 | | |
| | 113 | 18 | 16 | 21 | 17 | 3 | 6 | 13 | 1 | 4 | 7 | 5 | 22 | 14 | 19 | 11 | 15 | 8 | 20 | 9 | 23 | 2 | 10 | 12 | | |
| | 347 | 4 | 16 | 14 | 23 | 1 | 10 | 22 | 15 | 11 | 5 | 2 | 18 | 21 | 8 | 13 | 17 | 7 | 3 | 19 | 6 | 20 | 9 | 12 | | |
| 22 | 282 | 8 | 13 | 1 | 9 | 14 | 19 | 4 | 22 | 15 | 12 | 6 | 11 | 21 | 18 | 3 | 17 | 20 | 10 | 2 | 16 | 5 | 7 | | | |
| | 2044 | 13 | 18 | 12 | 5 | 8 | 1 | 11 | 16 | 2 | 6 | 21 | 9 | 15 | 19 | 22 | 3 | 17 | 4 | 7 | 10 | 20 | 14 | | | |
| | 1082 | 18 | 3 | 12 | 2 | 16 | 5 | 11 | 1 | 15 | 20 | 6 | 4 | 9 | 19 | 17 | 10 | 14 | 7 | 22 | 8 | 21 | 13 | | | |
| 21 | 404 | 6 | 11 | 16 | 10 | 15 | 21 | 3 | 1 | 9 | 14 | 5 | 19 | 2 | 12 | 18 | 7 | 4 | 17 | 20 | 8 | 13 | | | | |
| | 807 | 7 | 19 | 13 | 16 | 12 | 5 | 3 | 1 | 6 | 2 | 11 | 20 | 14 | 17 | 4 | 18 | 21 | 8 | 10 | 15 | 9 | | | | |
| | 592 | 16 | 5 | 13 | 21 | 7 | 18 | 11 | 3 | 1 | 4 | 10 | 17 | 19 | 12 | 15 | 6 | 14 | 9 | 20 | 8 | 2 | | | | |
| 20 | 453 | 12 | 7 | 11 | 3 | 20 | 18 | 14 | 4 | 6 | 8 | 5 | 15 | 17 | 9 | 2 | 16 | 19 | 10 | 1 | 13 | | | | | |
| | 279 | 9 | 17 | 1 | 10 | 2 | 16 | 18 | 8 | 12 | 3 | 15 | 6 | 20 | 13 | 5 | 7 | 19 | 14 | 11 | 4 | | | | | |
| | 170 | 4 | 16 | 9 | 6 | 1 | 17 | 20 | 5 | 13 | 11 | 18 | 2 | 7 | 12 | 15 | 3 | 8 | 10 | 14 | 19 | | | | | |
| 19 | 3166 | 1 | 5 | 9 | 13 | 17 | 8 | 12 | 15 | 6 | 3 | 19 | 4 | 14 | 18 | 10 | 2 | 11 | 16 | 7 | | | | | | |
| | 1462 | 2 | 15 | 9 | 11 | 13 | 8 | 1 | 3 | 19 | 10 | 14 | 17 | 6 | 18 | 12 | 5 | 16 | 4 | 7 | | | | | | |
| | 382 | 8 | 5 | 2 | 13 | 9 | 16 | 12 | 19 | 17 | 6 | 4 | 1 | 7 | 14 | 10 | 18 | 15 | 3 | 11 | | | | | | |
| 18 | 5730 | 12 | 3 | 11 | 13 | 5 | 9 | 1 | 4 | 7 | 16 | 10 | 17 | 15 | 18 | 8 | 2 | 14 | 6 | | | | | | | |
| | 271 | 13 | 4 | 16 | 1 | 10 | 7 | 3 | 12 | 17 | 2 | 6 | 18 | 11 | 14 | 8 | 15 | 5 | 9 | | | | | | | |
| | 1828 | 3 | 11 | 13 | 5 | 1 | 18 | 14 | 2 | 8 | 15 | 9 | 7 | 17 | 4 | 12 | 16 | 6 | 10 | | | | | | | |
| 17 | 1244 | 11 | 5 | 3 | 6 | 13 | 10 | 14 | 17 | 15 | 4 | 2 | 7 | 9 | 12 | 8 | 1 | 16 | | | | | | | | |
| | 338 | 15 | 2 | 12 | 17 | 8 | 5 | 11 | 9 | 3 | 16 | 13 | 10 | 1 | 6 | 4 | 7 | 14 | | | | | | | | |
| | 114 | 16 | 10 | 7 | 5 | 8 | 12 | 14 | 17 | 2 | 4 | 13 | 11 | 9 | 6 | 1 | 3 | 15 | | | | | | | | |
| 16 | 98 | 5 | 15 | 2 | 11 | 6 | 16 | 3 | 10 | 7 | 4 | 14 | 1 | 13 | 9 | 12 | 8 | | | | | | | | | |
| | 330 | 14 | 10 | 7 | 4 | 12 | 1 | 9 | 5 | 2 | 15 | 3 | 8 | 11 | 13 | 16 | 6 | | | | | | | | | |
| | 255 | 3 | 13 | 11 | 8 | 4 | 1 | 12 | 15 | 2 | 16 | 9 | 6 | 14 | 10 | 7 | 5 | | | | | | | | | |
| 15 | 378 | 12 | 8 | 5 | 13 | 6 | 10 | 2 | 4 | 14 | 9 | 11 | 15 | 7 | 1 | 3 | | | | | | | | | | |
| | 25 | 11 | 15 | 6 | 12 | 5 | 7 | 1 | 13 | 2 | 14 | 8 | 3 | 9 | 4 | 10 | | | | | | | | | | |
| | 2749 | 3 | 10 | 6 | 2 | 5 | 15 | 13 | 9 | 4 | 14 | 7 | 11 | 1 | 8 | 12 | | | | | | | | | | |
| 14 | 49 | 11 | 1 | 10 | 5 | 9 | 2 | 12 | 14 | 7 | 13 | 4 | 6 | 8 | 3 | | | | | | | | | | | |
| | 1077 | 1 | 14 | 7 | 2 | 12 | 9 | 6 | 3 | 10 | 4 | 13 | 8 | 5 | 11 | | | | | | | | | | | |
| | 31 | 2 | 6 | 11 | 9 | 7 | 1 | 4 | 14 | 12 | 8 | 5 | 3 | 13 | 10 | | | | | | | | | | | |
| 13 | 2272 | 12 | 6 | 3 | 10 | 4 | 11 | 8 | 2 | 7 | 13 | 1 | 9 | 5 | | | | | | | | | | | | |
| | 224 | 9 | 12 | 8 | 3 | 1 | 13 | 5 | 10 | 6 | 11 | 2 | 4 | 7 | | | | | | | | | | | | |
| | 8 | 8 | 13 | 3 | 6 | 11 | 5 | 12 | 9 | 4 | 2 | 7 | 10 | 1 | | | | | | | | | | | | |
| 12 | 2865 | 6 | 2 | 5 | 10 | 12 | 4 | 8 | 11 | 3 | 1 | 7 | 9 | | | | | | | | | | | | | |
| | 82 | 2 | 8 | 5 | 9 | 1 | 6 | 11 | 3 | 12 | 7 | 4 | 10 | | | | | | | | | | | | | |
| | 52 | 4 | 9 | 1 | 3 | 11 | 8 | 12 | 2 | 6 | 10 | 7 | 5 | | | | | | | | | | | | | |
| 11 | 320 | 10 | 8 | 6 | 3 | 9 | 11 | 1 | 5 | 7 | 2 | 4 | | | | | | | | | | | | | | |
| | 339 | 8 | 10 | 2 | 4 | 9 | 7 | 3 | 11 | 6 | 1 | 5 | | | | | | | | | | | | | | |
| | 97 | 3 | 9 | 6 | 4 | 11 | 1 | 8 | 2 | 5 | 7 | 10 | | | | | | | | | | | | | | |
| 10 | 2383 | 6 | 8 | 1 | 4 | 9 | 5 | 2 | 10 | 3 | 7 | | | | | | | | | | | | | | | |
| | 125 | 4 | 10 | 7 | 2 | 6 | 3 | 1 | 8 | 5 | 9 | | | | | | | | | | | | | | | |
| | 251 | 5 | 7 | 4 | 1 | 8 | 2 | 9 | 6 | 3 | 10 | | | | | | | | | | | | | | | |

## 5. Conclusion and future works:

This paper reveals that the *N*-Queens problem can be solved within a reasonable time by using metaheuristics like Genetic Algorithm and its modified forms. *N*-Queens Problems have a modicum of practical usage yet they represent a major class of NP problems that cannot be solved in a reasonable amount of time using deterministic methods- the reason of their importance. Genetic Algorithms were framed as heuristics for solving problems with good and bad solutions depending on the fitness values of the solutions, yet here they proved their ability to solve combinatorial optimization problems with simple binary (yes and no) answers. In the previous literatures, attempts were made to solve *N*-Queens Problems with Genetic Algorithms. In this paper with modifications in the fitness functions, hereby giving rise to a novel modified GA, it is seen that this approach yields promising and satisfactory results in less time compared to that obtained from the previous approaches.

The future works include the use of various other metaheuristics algorithms for solving the *N*-Queens Problem with more efficiency. Also application of the proposed algorithm and other variants of it in solving other scheduling problems will lead to interesting results.

## Appendix-A:

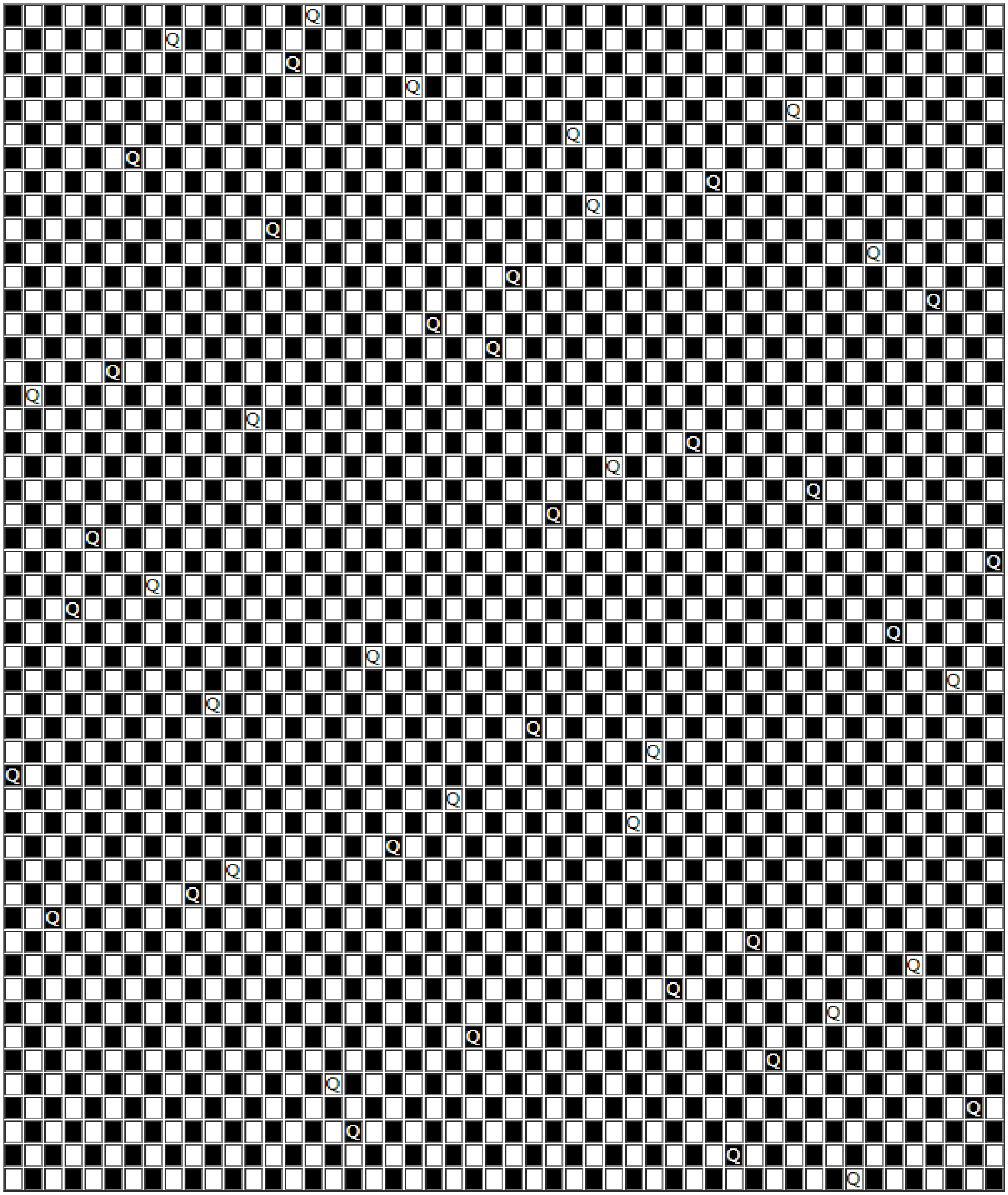

**Fig.4: An optimal Solution for 50x50 Chessboard**

Solution in Tuple: (18, 34, 12, 25, 28, 35, 44, 26, 49, 13, 21, 14, 33, 41, 48, 50, 5, 3, 23, 15, 47, 37, 17, 7, 36, 39, 20, 29, 45, 42, 31, 16, 19, 9, 32, 43, 2, 11, 6, 46, 30, 8, 1, 40, 24, 10, 38, 22, 4, 27)

## Appendix-B:

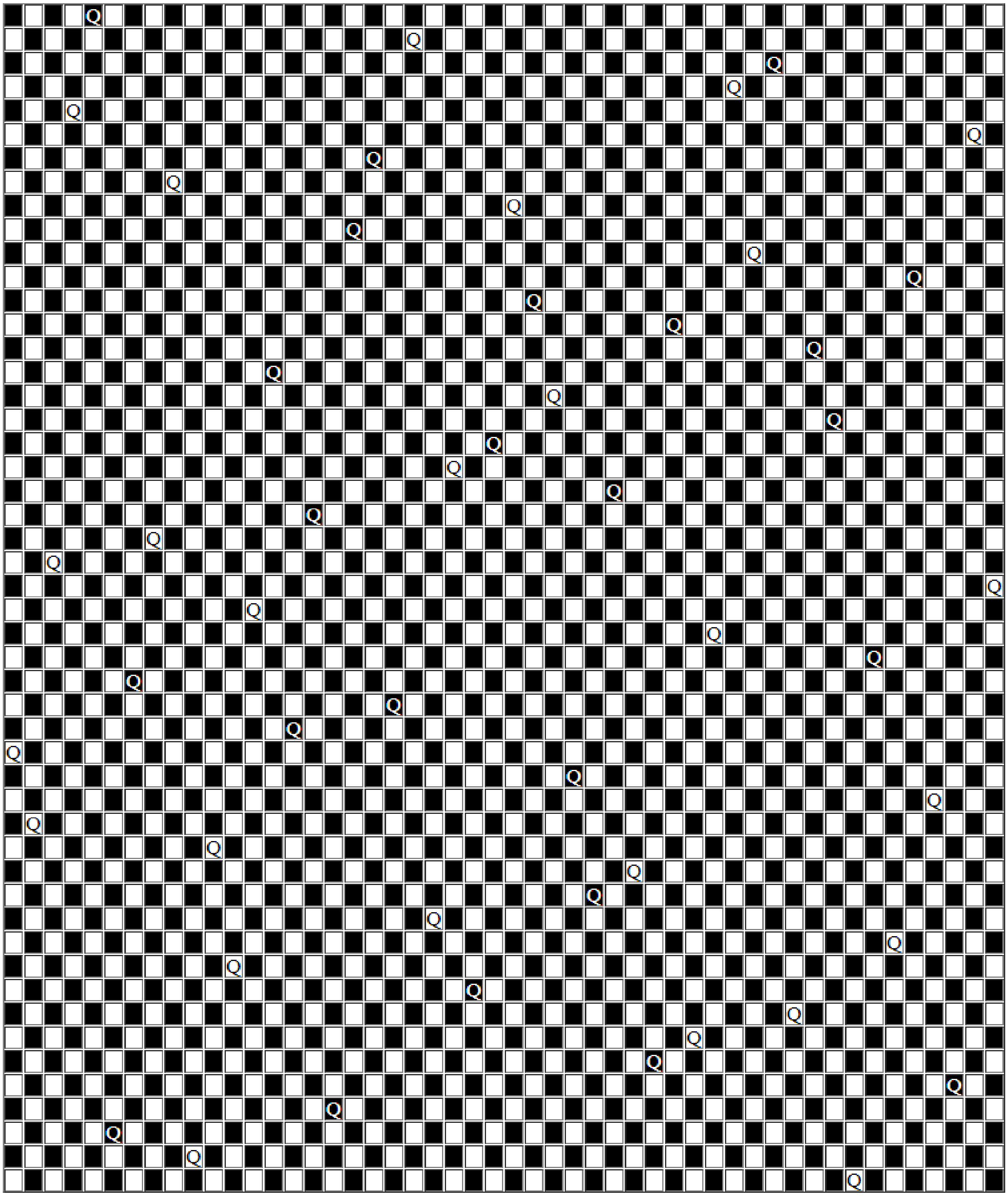

Fig.5: Another optimal Solution for 50x50 Chessboard

Solution in Tuple: (19, 16, 27, 46, 50, 3, 22, 28, 43, 2, 15, 10, 25, 35, 20, 29, 4, 41, 44, 21, 49, 12, 31, 9, 32, 42, 38, 34, 18, 13, 30, 14, 6, 37, 7, 24, 47, 40, 48, 8, 36, 33, 1, 23, 11, 39, 17, 5, 45, 26)

## Appendix-C:

Solution of 100-Quuens Problem:

| | | | | | | | | | | | |
|---|---|---|---|---|---|---|---|---|---|---|---|
| (34, | 27, | 61, | 99, | 2, | 80, | 66, | 11, | 54, | 47, | 51, | 85, |
| 83, | 60, | 45, | 7, | 37, | 41, | 18, | 70, | 17, | 87, | 10, | 48, |
| 79, | 75, | 88, | 90, | 98, | 1, | 63, | 24, | 42, | 44, | 91, | 22, |
| 100, | 49, | 28, | 5, | 8, | 57, | 65, | 73, | 76, | 36, | 21, | 31, |
| 84, | 52, | 59, | 29, | 94, | 23, | 81, | 95, | 50, | 30, | 6, | 96, |
| 39, | 43, | 69, | 77, | 92, | 25, | 72, | 3, | 97, | 19, | 14, | 86, |
| 71, | 56, | 38, | 16, | 78, | 82, | 4, | 26, | 12, | 89, | 15, | 35, |
| 40, | 9, | 53, | 64, | 74, | 58, | 32, | 20, | 68, | 55, | 33, | 46, |
| 67, | 13, | 93, | 62) | | | | | | | | |